\documentclass{llncs}
\usepackage{amsmath,graphicx,booktabs}
\usepackage{multirow}
\usepackage{epstopdf,url}
\usepackage{makeidx,amssymb}  
\usepackage{caption}
\usepackage{comment}
\usepackage{color}
\usepackage[caption=false,font=footnotesize]{subfig}

\newcommand{\ve}[1]{\mathbf{#1}} 

\begin{document}
	
\title{Automatic 3D Cardiovascular MR Segmentation with Densely-Connected Volumetric ConvNets}
\author{Lequan Yu\inst{1}, Jie-Zhi Cheng\inst{2}, Qi Dou\inst{1}, Xin Yang\inst{1}, Hao Chen\inst{1}, \\Jing Qin\inst{3}, \and Pheng-Ann Heng\inst{1,4} }
\institute{Dept. of Computer Science and Engineering, The Chinese University of Hong Kong
	\and Department of Electrical Engineering, Chang Gung University, Taoyuan, Taiwan
	\and Centre for Smart Health, School of Nursing, The Hong Kong Polytechnic University
	\and Guangdong Provincial Key Laboratory of Computer Vision and Virtual Reality Technology,
	Shenzhen Institutes of Advanced Technology, Chinese Academy of Sciences, Shenzhen, China}
\maketitle

\begin{abstract}
	Automatic and accurate whole-heart and great vessel segmentation from 3D cardiac magnetic resonance (MR) images plays an important role in the computer-assisted diagnosis and treatment of cardiovascular disease. However, this task is very challenging due to ambiguous cardiac borders and large anatomical variations among different subjects.
	In this paper, we propose a novel densely-connected volumetric convolutional neural network, referred as \emph{DenseVoxNet}, to automatically segment the cardiac and vascular structures from 3D cardiac MR images. The DenseVoxNet adopts the 3D fully convolutional architecture for effective volume-to-volume prediction. 
	From the learning perspective, our DenseVoxNet has three compelling advantages. First, it preserves the maximum information flow between layers by a densely-connected mechanism and hence eases the network training. Second, it avoids learning redundant feature maps by encouraging feature reuse and hence requires fewer parameters to achieve high performance, which is essential for medical applications with limited training data. Third, we add auxiliary side paths to strengthen the gradient propagation and stabilize the learning process.
	We demonstrate the effectiveness of DenseVoxNet by comparing it with the state-of-the-art approaches from HVSMR 2016 challenge in conjunction with MICCAI, and our network achieves the best dice coefficient. We also show that our network can achieve better performance than other 3D ConvNets but with fewer parameters.
\end{abstract}

\section{Introduction}
Accurate segmentation of cardiac structures in 3D cardiac MR images is crucial for the diagnosis and treatment planning of cardiovascular disease. For example, the segmentation results can support the building of patient-specific 3D heart model for the surgical planning of the severe congenital heart disease~\cite{pace2015interactive}. 
The manual segmentation on every MR slice can be very tedious and time-consuming, and subjects to inter- and intra-observer variability. Accordingly, an automatic segmentation scheme is highly demanded in clinical practice.

However, the automatic segmentation is by no means a trivial task, as some parts of cardiac borders are not very well defined due to the low contrast to the surrounding tissues. Meanwhile, the inter-subject variation of cardiac structures may impose more difficulty for the segmentation task. One prominent family of approaches are based on multiple atlases and deformable models~\cite{zhuang2013challenges}. These approaches needs to well consider the high anatomical variations in different subjects and useful atlases need to be built from a relatively large dataset. Pace et al.~\cite{pace2015interactive} developed an interactive method for the accurate segmentation of cardiac chambers and vessels, but this method is very slow. Recently, convolutional neural networks (ConvNets) significantly improve the segmentation performance for medical images~\cite{cciccek20163d,dou20163d,ronneberger2015u}. 
As for this task, Wolterink et al.~\cite{wolterink2017} employed a dilated ConvNet to demarcate the myocardium and blood pool, but the 3D volumetric information was not fully used in the study. Yu et al.~\cite{yu2017fractalnet} proposed the 3D FractalNet to consider the 3D image information. However, this network and other 3D ConvNets (e.g., 3D U-Net~\cite{cciccek20163d}, VoxResNet~\cite{chen2016voxresnet}) usually generate a large number of feature channels in each layer and they have plenty of parameters to be tuned during training. Although these networks introduce different skip connections to ease the training, the training of an effective model with the limited MR images for heart segmentation is still very challenging. 

In order to ease the training of 3D ConvNets with limited data, we propose a novel densely-connected volumetric ConvNet, namely \emph{DenseVoxNet}, to segment the cardiac and vascular structures in cardiac MR images. The DenseVoxNet adopts 3D fully convolutional architecture, and thus can fully incorporate the 3D image and geometric cues for effective volume-to-volume prediction. More importantly, the DenseVoxNet incorporates the concept of dense connectivity~\cite{huang2016densely} and enjoys three advantages from the learning perspective.
First, it implements direct connections from a layer to all its subsequent layers. Each layer can receive additional supervision from the loss function through the shorter connections, and thus make the network much easier to train. Second, the DenseVoxNet has fewer parameters than the other 3D ConvNets. Since layers can access to feature maps from all of its preceding layers, the learning of redundant feature maps can be possibly avoided. Therefore, the DenseVoxNet has fewer feature maps in each layer, which is essential for training ConvNets with limited images as it has less chance to encounter the overfitting problem. 
Third, we further improve the gradient flow within the network and stabilize the learning process via auxiliary side paths.
We extensively evaluate the DenseVoxNet on the HVSMR 2016 challenge dataset. The results demonstrate that DenseVoxNet can outperform other state-of-the-art methods for the segmentation of myocardium and blood pool in 3D cardiac MR images, corroborating its advantages over existing methods.

\section{Method}
\label{sec:method}
In this section, we first introduce the concept of dense connection. Then, we elaborate the architecture of our DenseVoxNet bearing the spirit of dense connection. The training procedure is detailed in the last subsection. 

\begin{figure}[t]
	\centering
	\includegraphics[width=0.95\linewidth]{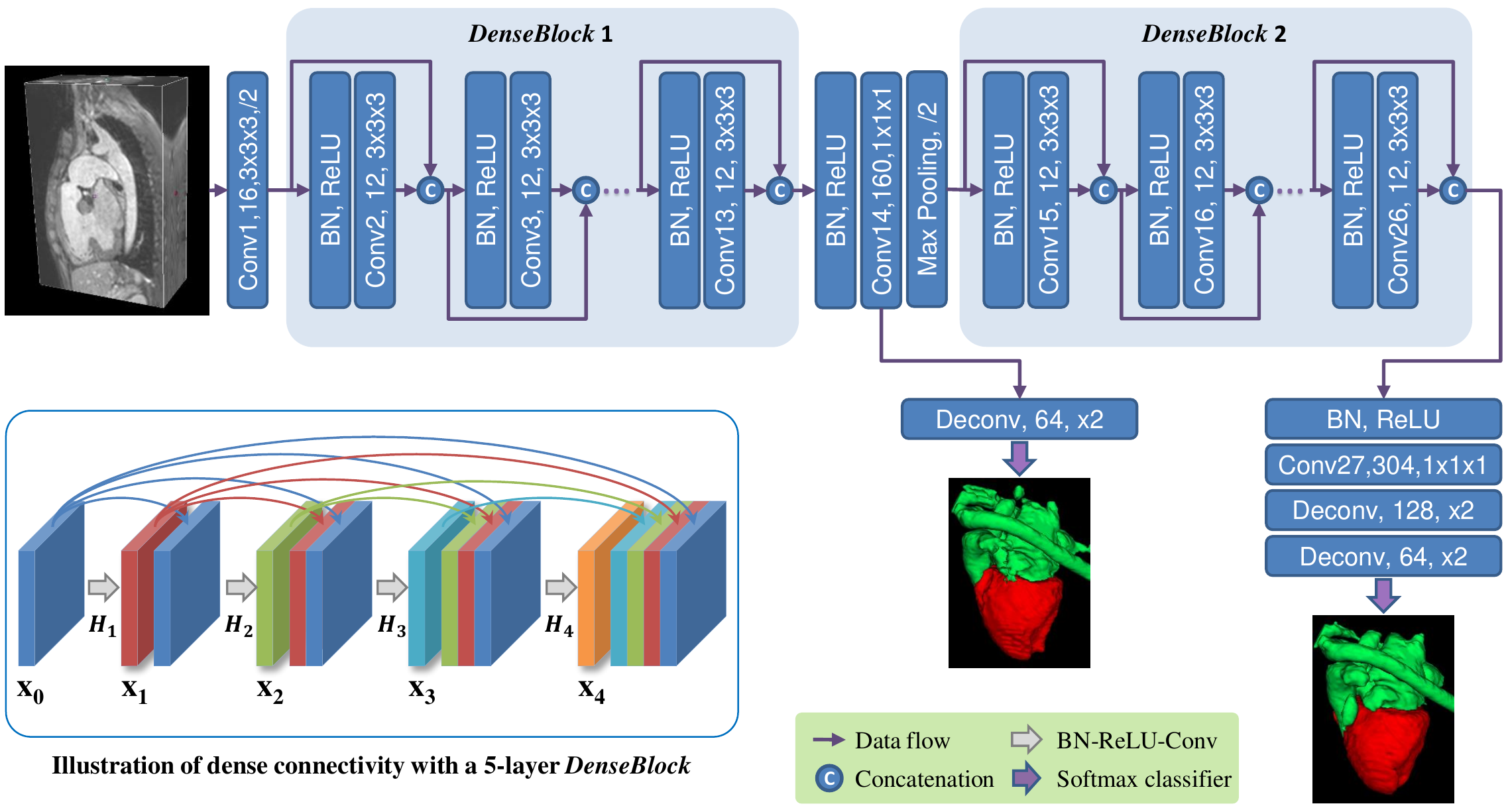}
	\caption{ The architecture of our DenseVoxNet. It consists of two \emph{DenseBlocks} and all operations are implemented in a 3D manner. The green and red color denotes the output of blood pool and myocardium. The graph in left bottom illustrates the dense connectivity scheme taking a 5-layer \emph{DenseBlock} as an example.}
	\label{fig:structure}
	\vspace{-0.3cm}
\end{figure}

\subsection{Dense Connection}
\label{densenet}
In a ConvNet, we denote $\ve{x}_{\ell}$ as the output of the $\ell^{th}$ layer, and $\ve{x}_{\ell}$ can be computed by a transformation $H_l(\ve{x})$ from the output of the previous layer, $\ve{x}_{\ell-1}$ as:
\begin{equation}
\ve{x}_\ell = H_\ell(\ve{x}_{\ell-1}),
\end{equation}
where $H_\ell(\ve{x})$ can be a composite of operations such as Convolution (Conv), Pooling, Batch Normalization (BN) or rectified linear unit (ReLU), etc.
To boost the training against the vanishing gradients, ResNet~\cite{he2016deep} introduces a kind of skip connection which integrates the response of $H_\ell(\ve{x})$ with the identity mapping of the features from the previous layer to augment the information propagation as:

\begin{equation}
\ve{x}_\ell = H_\ell(\ve{x}_{\ell-1}) + \ve{x}_{\ell-1}.
\end{equation}
However, the identity function and the output of $H_\ell$ are combined by summation, which may impede the information flow in the network. 

To further improve the information flow within the network, the dense connectivity~\cite{huang2016densely}  exercises the idea of skip connections to the extreme by implementing the connections from a layer to all its subsequent layers. Specifically, the $\ve{x}_\ell$ is defined as: 
\begin{equation}
\label{dense}
\ve{x}_\ell = H_\ell([\ve{x}_0, \ve{x}_1,...,\ve{x}_{\ell-1}]),
\end{equation}
where  $[...]$ refers to the concatenation operation. The dense connectivity, as illustrated at the left bottom of Fig.~\ref{fig:structure}, makes all layers receive direct supervision signal. 
More importantly, such a mechanism can encourage the reuse of features among all these connected layers. Suppose that if the output of each layer has $k$ feature maps, then the $k$, referred as \emph{growth rate}, can be set to a small number to reduce the number of parameters since there is no need to re-learn redundant feature maps.
This characteristic is quite compelling to medical image analysis tasks, where it is usually difficult to train an effective network with a lot of parameters with limited training data.

\subsection{The Architecture of DenseVoxNet}
Fig.~\ref{fig:structure} illustrates the architecture of our proposed DenseVoxNet. 
It adopts the 3D fully convolutional network architecture~\cite{chen2016voxresnet,cciccek20163d,dou20163d} and has the down- and up-sampling components to achieve end-to-end training. Note that the Eq.~\ref{dense} is not applicable when the feature maps have different sizes; on the another hand, we need to reduce the feature map size for better efficiency of memory space and increase the receptive field to enclose more information when prediction. We, therefore, divide the down-sampling components into two densely-connected blocks, referred as \emph{DenseBlock}, and each \emph{DenseBlock} is comprised of 12 transformation layers with dense connections (Only draw 3 layers in the figure for simplicity). Each transformation layer is sequentially composed of a BN, a ReLU, and a 3$\times$3$\times$3 Conv and the growth rate, $k$, of our DenseVoxNet is 12. The first \emph{DenseBlock} is prefixed with a Conv with 16 output channels and stride of 2 to learn primitive features.
In-between the two \emph{DenseBlocks} is the transition block which consists of a BN, a ReLU, a 1$\times$1$\times$1 Conv and a 2$\times$2$\times$2 max pooling layers. 

The up-sampling component is composed of a BN, a ReLU, a 1$\times$1$\times$1 Conv and two 2$\times$2$\times$2 deconvolutional (Deconv) layers to ensure the sizes of segmentation prediction map consistent with the size of input images. 
The up-sampling component is then followed with a 1$\times$1$\times$1 Conv layer and soft-max layer to generate the final label map of the segmentation. To equip the DenseVoxNet with the robustness against the overfitting problem, the dropout layer is implemented following each Conv layer with the dropout rate of 0.2.

To further boost the information flow within the network, we implement a kind of long skip connection to connect the transition layer to the output layer with a 2$\times$2$\times$2 Deconv layer. This skip connection shares the similar idea of deep supervision~\cite{dou20163d} to strengthen the gradient propagation and stabilize the learning process. In addition, this long skip connection may further tap the potential of the limited training data to learn more discriminative features. Our DenseVoxNet has about 1.8M parameters in total, which is much fewer than 3D U-Net~\cite{cciccek20163d} with 19.0M parameters and VoxResNet~\cite{chen2016voxresnet} with 4.0M parameters. 

\subsection{Training Procedure}
The DenseVoxNet is implemented with Caffe~\cite{jia2014caffe} library\footnote{\url{https://github.com/yulequan/HeartSeg}}. The weights were randomly initialized with a Gaussian distribution ($\mu$ = 0, $\sigma$ = 0.01). The optimization is realized with the stochastic gradient descend algorithm (batch size = 3, weight decay = 0.0005, momentum = 0.9). The initial learning rate was set to 0.05. We use the ``poly" learning rate policy (i.e., the learning rate is multiplied by $(1-\frac{iter}{max\_iter})^{power}$) for the decay of learning rate along the training iteration. The power variable was set to 0.9 and maximum iteration number (max\_iter) was set as 15000. 
To fit the limited 12GB GPU memory, the input of the DenseVoxNet is sub-volumes with size of 64$\times$64$\times$64, which were randomly cropped from the training images. The final segmentation results were obtained with the major voting strategy~\cite{kontschieder2011structured} from the predictions of the overlapped sub-volumes.

\section{Experiments and Results}
\subsubsection{Dataset and Pre-processing}
The DenseVoxNet is evaluated with the dataset of HVSMR 2016 Challenge. There are in total 10 3D cardiac MR scans for training and 10 scans for testing. The scans have low quality as they were acquired with a 1.5T scanner. 
All cardiac MR images were scanned from the patients with congenital heart diseases (CHD). The HVSMR 2016 dataset contains the annotations for the myocardium and great vessel, and the testing data annotations are held by organizers for fair comparison.
Due to the large intensity variance among different images, all cardiac MR images were normalized to have zero mean and unit variance. 
We did not employ spatial resampling. 
To leverage the limited training data, simple data augmentation was employed to enlarge the training data. The augmentation operations include the rotation with 90, 180 and 270 degrees, as well as image flipping along the axial plane.

\subsubsection{Qualitative Results} 
\begin{figure}[t]
	\centering
	\includegraphics[width=0.90\linewidth]{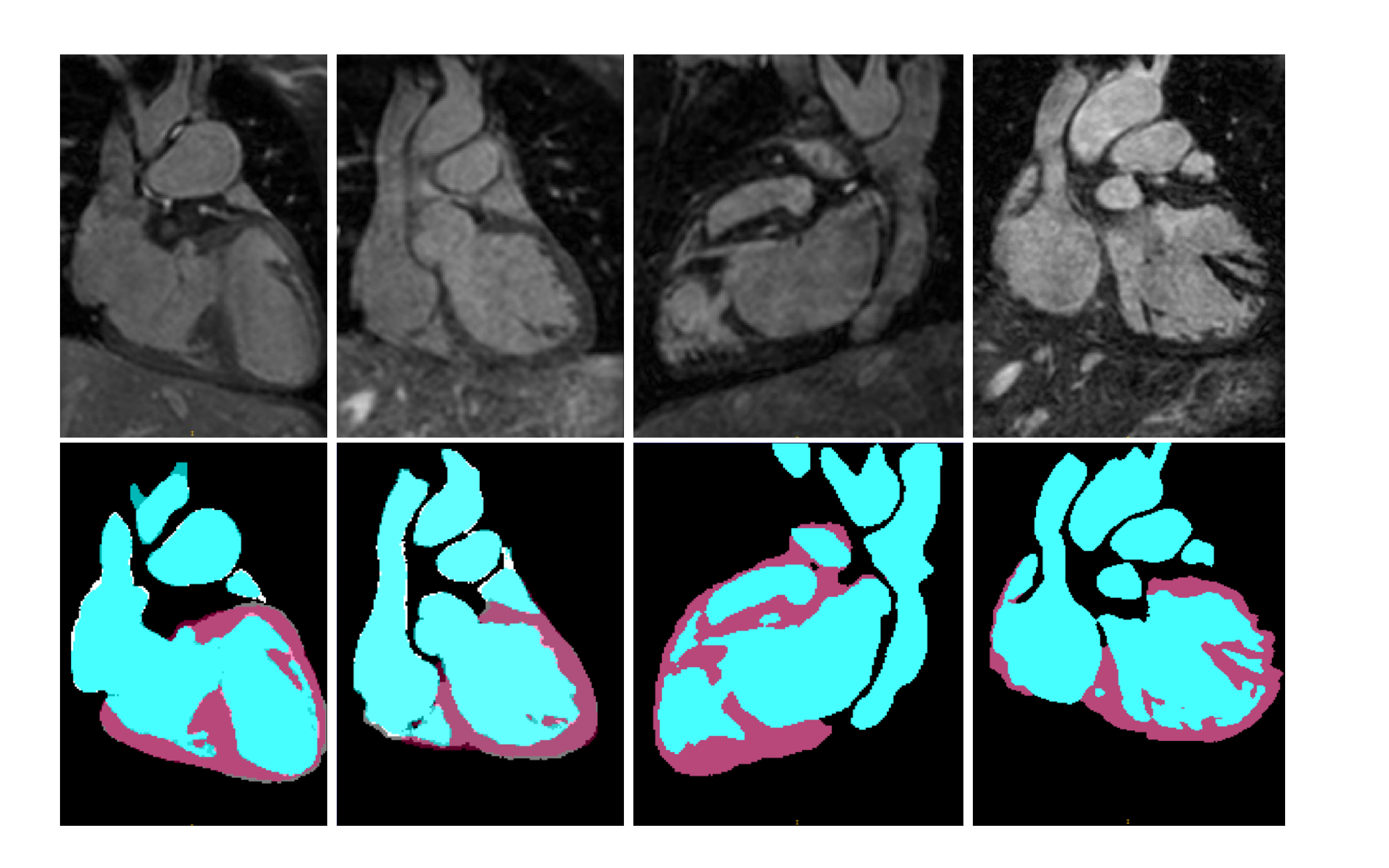}
	\caption{Segmentation results on training images (the first two) and testing images (the last two). The blue and purple color denotes our segmentation results for blood pool and myocardium, respectively, and segmentation ground truth is also presented in white and gray regions in the first two samples.}
	\label{fig:qualitative_show}
	\vspace{-0.3cm}
\end{figure}

In Fig.~\ref{fig:qualitative_show}, we demonstrate 4 typical segmentation results on training images (the first two samples, via cross validation) and testing images (the last two samples). The four slices are from different subjects but with the same coronal plane view. The blue and purple color denotes our segmentation results for blood pool and myocardium, respectively, and segmentation ground truth is also presented in white and gray regions in the first two samples. As can be observed, there exists large variation of cardiac structures among different subjects in both training and testing images. Our method can still successfully demarcate myocardium and blood pool from the low-intensity contrast cardiac MR images, demonstrating the effectiveness of the proposed DenseVoxNet.
\subsubsection{Comparison with Other Methods}
The quantitative comparison between DenseVoxNet and other approaches from the participating teams in this challenge is shown in Table~\ref{tab:results}. According to the rules of the challenge, methods were ranked based on Dice coefficient (Dice). Meanwhile, other ancillary measures like average surface distance (ABD) and symmetric Hausdorff distance (Hausdorff) are also computed for reference. Higher Dice values suggest a higher agreement between segmentation results and ground truth, while lower ABD and Hausdorff values indicate higher boundary similarity. 
Three of the six approaches employed traditional methods based on hand-crafted features, including random forest~\cite{Mukhopadhyay2017}, 3D Markov random field and substructure tracking~\cite{Tziritas2017} and level-set method driven by multiple atlases~\cite{van2017automated}. The other three methods, include ours, are based on ConvNet. Wolterink et al.~\cite{wolterink2017} employed 2D dilated ConvNets to segment the myocardium and blood pool, while Yu et al.~\cite{yu2017fractalnet} utilized 3D ConvNets.

\begin{table*}[]
	\vspace{-0.7cm}
	\centering
	\caption{Comparison with different approaches on HVSMR2016 dataset.}
	\label{tab:results}
	\resizebox{1.0\textwidth}{!}{
		\begin{tabular}{ c| c | c | c || c | c | c}
			\hline
			\multirow{2}{*}{Method} & \multicolumn{3}{c||}{Myocardium} & \multicolumn{3}{c}{Blood pool} \\
			\cline{2-7} 		& Dice	& ADB [mm]	& Hausdorff [mm]	& Dice	& ADB [mm]	& Hausdorff [mm] \\ \hline
			Mukhopadhyay~\cite{Mukhopadhyay2017} 		&0.495$\pm$0.126 &2.596$\pm$1.358 &12.796$\pm$4.435&0.794$\pm$0.053 &2.550$\pm$0.996 &14.634$\pm$8.200 \\ \hline
			Tziritas~\cite{Tziritas2017}				&0.612$\pm$0.153 &2.041$\pm$1.022 &13.199$\pm$6.025&0.867$\pm$0.047 &2.157$\pm$0.503 &19.723$\pm$4.078 \\ \hline
			Shahzad et al.~\cite{van2017automated}		&0.747$\pm$0.075 &1.099$\pm$0.204 &\textbf{5.091$\pm$1.658} &0.885$\pm$0.028 &1.553$\pm$0.376 &9.408$\pm$3.059 \\ \hline	
			Wolterink et al.~\cite{wolterink2017} &0.802$\pm$0.060 &\textbf{0.957$\pm$0.302} &6.126$\pm$3.565 &0.926$\pm$0.018 &0.885$\pm$0.223 &7.069$\pm$2.857 \\ \hline
			Yu et al.~\cite{yu2017fractalnet}			&0.786$\pm$0.064 &0.997$\pm$0.353 &6.419$\pm$2.574 &0.931$\pm$0.016 &\textbf{0.868$\pm$0.218} &\textbf{7.013$\pm$3.269} \\ \hline
			DenseVoxNet	(Ours)							&\textbf{0.821$\pm$0.041} &0.964$\pm$0.292 &7.294$\pm$3.340 &\textbf{0.931$\pm$0.011} &0.938$\pm$0.224 &9.533$\pm$4.194 \\ \hline
		\end{tabular}		
	}
	\vspace{-0.5cm}
\end{table*}

Table~\ref{tab:results} reports the results of different methods.
It can be observed that the ConvNet-based methods (the last three rows) can generally achieve better performance than the other methods do, suggesting that ConvNets can generate more discriminative features in a data-driven manner to better tackle the large anatomical variability of patients with CHD.
Regarding the segmentation of myocardium, our method achieves the best performance with the Dice, i.e., the ranking metric in the Challenge, of 0.821$\pm$0.041 and outperforms the second one by around $2\%$. For the segmentation of blood pool, our method also achieves the best Dice score of 0.931$\pm$0.011 with a small deviation.  The ADB and Hausdorff scores of our method are also competitive compared to the best performance. 
It is worth noting that the dice scores of myocardium in all methods are lower than the Dice scores of blood pool, suggesting that the segmentation of myocardium is relatively more challenging due to the ambiguous borders of the myocardium in the low-resolution MR images.
While other two ConvNet-based approaches achieve quite close Dice scores to our DenseVoxNet in blood pool segmentation, our method is obviously better than these two methods in the dice scores of the myocardium, demonstrating our densely-connected network with auxiliary long side paths has the capability to tackle hard myocardium segmentation problem.  

We further implement other two state-of-the-art 3D ConvNets, 3D U-Net~\cite{cciccek20163d} and VoxResNet~\cite{chen2016voxresnet}, for comparison. We also compare the performance of the proposed DenseVoxNet with and without auxiliary side paths. 
The quantitative comparison can be found in Table~\ref{tab:othernetworkresults}, where ``DenseVoxNet-A'' denotes the DenseVoxNet without the auxiliary side paths.
As can be observed, our DenseVoxNet achieves much better performance than the other two 3D ConvNets in both myocardium and blood pool segmentation. It suggests that our DenseVoxNet can benefit from the improved information flow throughout the network with the dense connections.
In addition, our method achieves better performance with much fewer parameters than our competitors, corroborating the effectiveness of the feature map reusing mechanism encoded in the densely-connected architecture, which is quite important to enhance the capability of ConvNet models under limited training data.
It is also observed that the auxiliary side path can further improve the segmentation performance, especially for the myocardium.
\begin{table*}[]
	\vspace{-0.7cm}
	\centering
	\caption{Quantitative analysis of our network}
	\label{tab:othernetworkresults}
	\resizebox{1.0\textwidth}{!}{
		\begin{tabular}{ c| c| c | c | c || c | c | c}
			\hline
			\multirow{2}{*}{Method} &\multirow{2}{*}{Parameters} & \multicolumn{3}{c||}{Myocardium} & \multicolumn{3}{c}{Blood pool} \\
			\cline{3-8} 		& & Dice	& ADB[mm]	& Hausdorff[mm]	& Dice	& ADB[mm]	& Hausdorff[mm] \\ \hline
			3D U-Net~\cite{cciccek20163d}	&19.0M &0.694$\pm$0.076 &1.461$\pm$0.397 &10.221$\pm$4.339&0.926$\pm$0.016 &0.940$\pm$0.192 &\textbf{8.628$\pm$3.390}\\ \hline
						
			VoxResNet~\cite{chen2016voxresnet}	  &4.0M &0.774$\pm$0.067 &1.026$\pm$0.400 &\textbf{6.572$\pm$3.551} &0.929$\pm$0.013 &0.981$\pm$0.186 &9.966$\pm$3.021\\ \hline
			
			DenseVoxNet-A				  &1.7M	&0.787$\pm$0.042 &1.811$\pm$0.752 &17.534$\pm$7.838 &0.917$\pm$0.018&1.451$\pm$0.537 &15.892$\pm$6.772 \\ \hline			
			
			DenseVoxNet	&1.8M  &\textbf{0.821$\pm$0.041} &\textbf{0.964$\pm$0.292} &7.294$\pm$3.340 &\textbf{0.931$\pm$0.011} &\textbf{0.938$\pm$0.224} &9.533$\pm$4.194 \\ \hline
		\end{tabular}
	}
	\vspace{-1.0cm}
\end{table*}

\section{Discussion and Conclusion}
A DenseVoxNet is proposed to automatically segment the cardiac structures in the 3D cardiac MR images. The DenseVoxNet is equipped with dense connectivity and spares network architecture from a large number of redundant features. It is because the learned features from previous layers can be reused. 
Therefore, the DenseVoxNet may enjoy better parameter efficiency and has less chance to encounter the overfitting problem when training with limited data.
We use lots of Conv layers in downsampling path and hence equip the network with large receptive fields to learn sufficient higher level features.
The denseVoxNet can attain best Dice scores for the segmentation of myocardium and blood pool on the challenge dataset. On the other hand, it is also interesting to observe that the 2D ConvNet method~\cite{wolterink2017} can outperform some 3D ConvNet methods on some metrics. 
It may be because the dataset in the HVSMR 2016 challenge is quite limited and it is very difficult to train an effective 3D network with such limited data. 
On the other hand, the DenseVoxNet can achieve better segmentation performance than the three 3D ConvNets do. Therefore, the efficacy of the DenseVoxNet can then be well corroborated. 
\\
\\
\textbf{Acknowledgments.}
The work described in this paper was supported by the grants from the Research Grants Council of the Hong Kong Special Administrative Region (Project No. CUHK 412513 and CUHK 14203115) and the National Natural Science Foundation of China (Project No. 61233012).

\bibliographystyle{splncs}
\bibliography{refs}
\end{document}